\documentclass[12pt]{article}
\usepackage{geometry,xcolor}
\usepackage{graphicx,longtable}
\usepackage{setspace}
\usepackage{lmodern}
\usepackage{microtype}
\usepackage{amsmath,amssymb,amsthm,mathtools}
\usepackage{needspace}
\usepackage{booktabs,tabularx,array}
\usepackage{placeins}
\usepackage[round]{natbib}
\usepackage{xr-hyper}
\usepackage{hyperref}
\hypersetup{
    colorlinks=true,
    linkcolor=blue,
    citecolor=blue,
    urlcolor=blue,
    filecolor=blue
}
\makeatletter
\newcommand*{\addFileDependency}[1]{%
  \typeout{(#1)}\@addtofilelist{#1}%
  \IfFileExists{#1}{}{\typeout{No file #1.}}}
\makeatother
\newcommand*{\myexternaldocument}[1]{%
  \externaldocument[][nocite]{#1}[#1.pdf]%
  \addFileDependency{#1.tex}\addFileDependency{#1.aux}}
\myexternaldocument{supporting}

\numberwithin{equation}{section}
\allowdisplaybreaks[2]

\newcommand{\E}{\mathbb{E}}
\newcommand{\Pp}{\mathbb{P}}
\newcommand{\ind}{\mathbf{1}}

\newcommand{\R}{\mathbb R}
\newcommand{\Hh}{\mathcal H}
\newcommand{\Ff}{\mathcal F}
\newcommand{\Dd}{\mathcal D}
\newcommand{\Risk}{\mathcal R}

\newcommand{\Eopt}{\mathcal E_{\mathrm{opt}}}

\newcommand{\norm}[1]{\left\lVert #1\right\rVert}

\newtheorem{theorem}{Theorem}[section]

\newtheorem{corollary}{Corollary}[section]

\theoremstyle{definition}
\newtheorem{example}{Example}
\theoremstyle{plain}

\title{Tabular Foundation Models for Weighted Regression and Classification}
\author{Junwen Xia\(^{1}\), Yan He\(^{2}\), Dehan Kong\(^{3}\), and Zhenhua Lin\(^{4*}\)\\
{\small {\it\(^1\)School of Statistics, Renmin University of China, Beijing, China}}\\
{\small {\it\(^2\)School of Statistics, East China Normal University, Shanghai, China}}\\
{\small {\it\(^3\)Department of Statistical Sciences, University of Toronto, Toronto, Ontario, Canada}}\\
{\small {\it\(^4\)Department of Statistics and Data Science, National University of Singapore, Singapore}}\\
{\small {\it\(^*\)email: linz@nus.edu.sg}}
}
\date{\today}

\begin{document}
{\singlespacing\maketitle}
\vspace{-2.2em}

\begin{abstract}
\onehalfspacing
Weighted regression and classification have applications ranging from missing-data analysis and causal inference to insurance pricing and particle physics. Existing weighted methods must be fitted and tuned on each dataset, which is costly for flexible nonlinear learners. Tabular foundation models avoid this cost. A Transformer is pretrained once on many simulated datasets and then predicts on a new dataset by in-context learning, using its observations as context without further fitting. Recent models such as TabPFN-3.5 surpass tuned conventional methods, but they cannot use sample weights. We develop a general pipeline that adapts such models to weighted prediction through weighted attention, a normalized weighted loss, and pretraining on synthetic weighted prediction tasks from missing-data problems. Applied to TabPFN-3.5, it yields WeightedPFN, which performs strongly on real datasets, both in missing-data problems and in out-of-distribution causal tasks, while running about as fast as linear baselines. We also establish a statistical theory showing that a suitably pretrained weighted Transformer attains the nonparametric minimax rate for weighted regression on new datasets without further fitting.
% The analysis links weighted attention to local polynomial estimation and provides constructions and error bounds that also apply to ordinary in-context regression.

\end{abstract}

\clearpage
{
\section{Introduction}
\label{sec:introduction}

Weighted regression and classification are fundamental tools that support a wide range of applications. For example, problems in survey sampling, missing-data analysis, distribution shift, causal inference, learning from unreliable observations, insurance claim modeling, and particle-physics simulation can be addressed by weighted regression and classification \citep{pfeffermann1993role,robins1995semiparametric,shimodaira2000,tian2014simple,zhao2012owl,ren2018learning,mccullagh1989,hayrapetyan2025reweighting}. We illustrate the applications of weighted regression and classification through two examples: missing-data prediction and treatment selection.

\begin{example}[Missing-data prediction]
\label{ex:intro-missing-data}
When covariates or responses are missing, the complete cases may differ systematically from the original population of interest before missingness. Under appropriate assumptions, weighting each complete case by the inverse of its observation probability can correct this selection bias. Predicting continuous or categorical responses then becomes a weighted regression or classification problem, respectively \citep{rubin1976inference,robins1995semiparametric}.
\end{example}

\begin{example}[Treatment selection]
\label{ex:intro-treatment-selection}
Treatment selection aims to recommend the treatment that yields the highest expected outcome for an individual. Under standard causal assumptions, this can be formulated either as weighted regression, which estimates the conditional average treatment effect (CATE) and assigns treatment according to its sign, or as weighted classification, which learns a treatment rule directly. The weights are constructed from the propensity score and outcome information \citep{tian2014simple,zhao2012owl}.
\end{example}

Formally, weighted regression and classification aim to learn a predictor $f$ that minimizes the weighted population risk $\E\{W\ell(f(X),Y)\}$, where $X$ denotes covariates, $Y$ the response, $W\geq 0$ a sample weight, and $\ell$ the squared-error loss for regression and the 0--1 loss for classification.
Conventional approaches include weighted generalized linear models \citep{mccullagh1989}, weighted support vector regression and classification \citep{cortes1995support,smola2004tutorial}, weighted tree ensembles \citep{breiman2001random,ke2017lightgbm}, and weighted neural networks \citep{rumelhart1986learning}. These methods trade flexibility against computational cost: weighted generalized linear models are fast to fit but may miss complex nonlinear relationships, whereas more flexible nonlinear methods require substantial time for training and hyperparameter tuning. Moreover, each method must be fitted and tuned separately for every new dataset, so this cost is incurred repeatedly. This motivates a reusable weighted learner that transfers knowledge across tasks, combining flexible weighted prediction with fast adaptation to new datasets without task-specific fitting or tuning.

Transformer-based in-context learning \citep{vaswani2017attention,hollmann2023tabpfn} offers a promising way to build such a learner. Pretrained tabular foundation models based on in-context learning can combine strong predictive performance with low computation time on unweighted prediction, using observed examples as context to predict responses for new observations without task-specific parameter updates or hyperparameter tuning. Recent models such as TabPFN-3.5 surpass extensively tuned conventional methods in aggregate performance on TabArena, a benchmark spanning diverse regression and classification datasets, while offering a favorable accuracy--runtime trade-off \citep{erickson2026tabarena,priorlabs2026tabpfn35}. However, existing tabular foundation models are designed for unweighted prediction and do not directly support sample weights. Adapting these models to weighted prediction requires a new pipeline for generating synthetic weighted regression and classification tasks, incorporating sample weights into the Transformer architecture, and training with weighted losses.

In this paper, we adapt tabular foundation models to weighted regression and classification. Our contributions are threefold.

First, we develop a reusable adaptation pipeline for weighted regression and classification. The pipeline has three components: synthetic weighted tasks for pretraining, a weighted attention mechanism that allows the Transformer to use sample weights, and a normalized weighted loss. We propose to generate diverse pretraining tasks exclusively from synthetic missing-data prediction (Example~\ref{ex:intro-missing-data}). These tasks are designed to teach the models how to use sample weights, so that they can transfer this capability to other weighted prediction problems without requiring each application to be included in pretraining. The pipeline is designed for reuse across attention-based tabular foundation models, so that advances in the underlying models can be incorporated into weighted learning.

Second, we apply this pipeline to TabPFN-3.5 \citep{priorlabs2026tabpfn35}, yielding \emph{WeightedPFN}, to our knowledge the first tabular foundation model specifically pretrained for weighted regression and classification\footnote{Shortly after we built our pipeline, the TabPFN GitHub repository introduced an experimental sample-weight interface, but according to its November 17, 2026 release notes, no released model supports this feature yet (\url{https://github.com/PriorLabs/TabPFN/releases/tag/v10.1.0}).}. We fine-tune the released parameters of TabPFN-3.5 for approximately half an hour on a single NVIDIA H100 GPU and obtain our WeightedPFN. Once pretrained, WeightedPFN makes weighted predictions on new datasets without task-specific fine-tuning or hyperparameter optimization. Experiments on 110 real regression datasets and 156 real classification datasets demonstrate strong performance in missing-data prediction (Example~\ref{ex:intro-missing-data}). WeightedPFN also generalizes well to out-of-distribution causal tasks (Example~\ref{ex:intro-treatment-selection}) built from 78 real randomized-experiment datasets. Across both task families, its runtimes are comparable to those of linear models and substantially below those of tuned nonlinear methods. The fine-tuned WeightedPFN models are available at \url{https://huggingface.co/xiajunwen/WeightedPFN}, and the inference code is available at \url{https://github.com/xiajunwen1007/WeightedPFN}.

Third, we establish a statistical theory of weighted in-context regression. Generalizing studies of Transformers as learning algorithms \citep{garg2022what,bai2023,kim2024transformers,ching2026}, we analyze a weighted Transformer that more closely reflects the architectures used in tabular foundation models than those in previous analyses by retaining softmax attention and Gaussian error linear unit (GELU) activations \citep{hendrycks2016}. Under smoothness, boundedness, and task-coverage assumptions, with suitable pretraining resources, the model attains the classical nonparametric minimax rate $n^{-2s/(2s+d)}$ for an $s$-smooth weighted regression function with $n$ context observations and $d$ covariates \citep{stone1982}. With all weights equal to one, the analysis covers ordinary in-context regression and thus provides theoretical support for why pretrained Transformers, such as those in unweighted tabular foundation models, can predict well on new datasets without task-specific fitting.

\subsection{Related work}
\label{sec:related-work}
Our work is most closely related to tabular foundation models for unweighted regression and classification. TabPFN pioneered tabular foundation models: a Transformer is pretrained on synthetic datasets and then makes predictions on real datasets in context \citep{hollmann2023tabpfn,hollmann2025tabpfn}. Subsequent work improves predictive performance, scalability, and computational efficiency. TabICL combines more diverse synthetic pretraining tasks, scalable attention, and improved optimization to achieve strong regression and classification performance \citep{qu2026tabiclv2}. LimiX uses context-conditional masked modeling to learn joint tabular structure, supporting regression, classification, and missing-value imputation within a unified framework \citep{zhang2026limix2}. TabSwift uses a lightweight row-wise attention backbone to provide efficient regression and classification while retaining competitive predictive performance \citep{liu2026tabswift}. The recent TabPFN-3.5 model achieves strong accuracy on the TabArena benchmark, outperforming existing tuned baselines across a broad range of regression and classification tasks \citep{erickson2026tabarena,priorlabs2026tabpfn35}.

Alongside these advances in tabular prediction, pretrained in-context models have been adapted to causal inference \citep{robertson2025dopfn}, time-series forecasting \citep{hoo2025tables}, survival analysis \citep{seletkov2026survival}, graph learning \citep{eremeev2026graphpfn}, and reinforcement learning \citep{schiff2026icrrl}. These extensions broaden the use of pretrained models to a wider range of statistical problems. However, none of these models can handle weighted regression or classification.

% Adapting pretrained learners to weighted objectives is closely related to work on class imbalance and reweighting. Existing studies adjust classification thresholds or downsample observations to address class imbalance in PFNs \citep{mcdowell2026imbalance}, or learn demonstration weights to improve language-model predictions \citep{yang2023reweighting}. Our focus is on observation weights supplied with the data to define the target population risk. We develop a common adaptation pipeline that enables tabular foundation models to use these weights directly in regression and classification.
\subsection{Outline}
This paper is organized as follows. Section~\ref{sec:methodology} formulates the weighted regression and classification problems and presents the synthetic task generation, weighted attention architecture, and normalized weighted loss. Section~\ref{sec:theory} develops the theoretical guarantees. Section~\ref{sec:experiments} describes the implementation and evaluates missing-data prediction and treatment selection on real datasets. We end with a discussion in Section~\ref{sec:discussion}. The supplementary material provides additional task constructions, architectural details, and proofs.
}

\section{Methodology}
\label{sec:methodology}
Consider a task with observations $(X,Y,W)$ drawn from a distribution \(P\), where $X\in\mathcal X\subseteq\R^d$ is a vector of $d$ covariates, {$Y$ is the response, taking values in $\mathbb{R}$ for regression or $[K]=\{1,\ldots,K\}$ for classification, and $W\geq 0$ denotes a sample weight.}

In conventional unweighted learning, the objective is to minimize the population risk $\E_P\{\ell(f(X),Y)\}$ over a class of prediction functions $f$, where $\ell$ denotes a loss function. The resulting optimal prediction depends on the choice of loss. {For regression, the squared-error loss \(\ell(a,b)=(a-b)^2\) for \(a,b\in\mathbb{R}\) yields the conditional mean $\E_P(Y\mid X=x)$ as the population risk minimizer. For multiclass classification, the 0--1 loss \(\ell(a,b)=\ind(a\neq b)\) for \(a,b\in[K]\) yields the optimal classification rule \(\arg\max_k \Pp_P(Y=k\mid X=x)\).}

In many applications, however, observations contribute unequally to the learning objective. {We therefore consider the weighted population risk $\E_P\{W\ell(f(X),Y)\}$.} Its minimizer over $f$ generally differs from its unweighted counterpart. Specifically, under squared-error loss, the optimal prediction is the weighted conditional mean $m_P(x)=\E_P(WY\mid X=x)/\E_P(W\mid X=x)$.
For multiclass classification, minimizing the weighted 0--1 loss yields the optimal classification rule
$\arg\max_k p_k^*(x)$, where $p_k^*(x)=\E_P[W\ind(Y=k)\mid X=x]/\E_P[W\mid X=x]$.

These formulas show why the weights matter: changing them can change the optimal prediction. A tabular foundation model that ignores the weights estimates the unweighted target, such as $\E_P(Y\mid X=x)$, rather than $m_P(x)$. We therefore build a weighted tabular foundation model that takes the weighted observations as context and estimates the weighted conditional mean or classification rule through in-context learning. A single model is pretrained across many synthetic datasets and then adapts to each new real dataset through its weighted context, without further fitting.

\begin{figure}[!htbp]
\centering
\includegraphics[width=\linewidth]{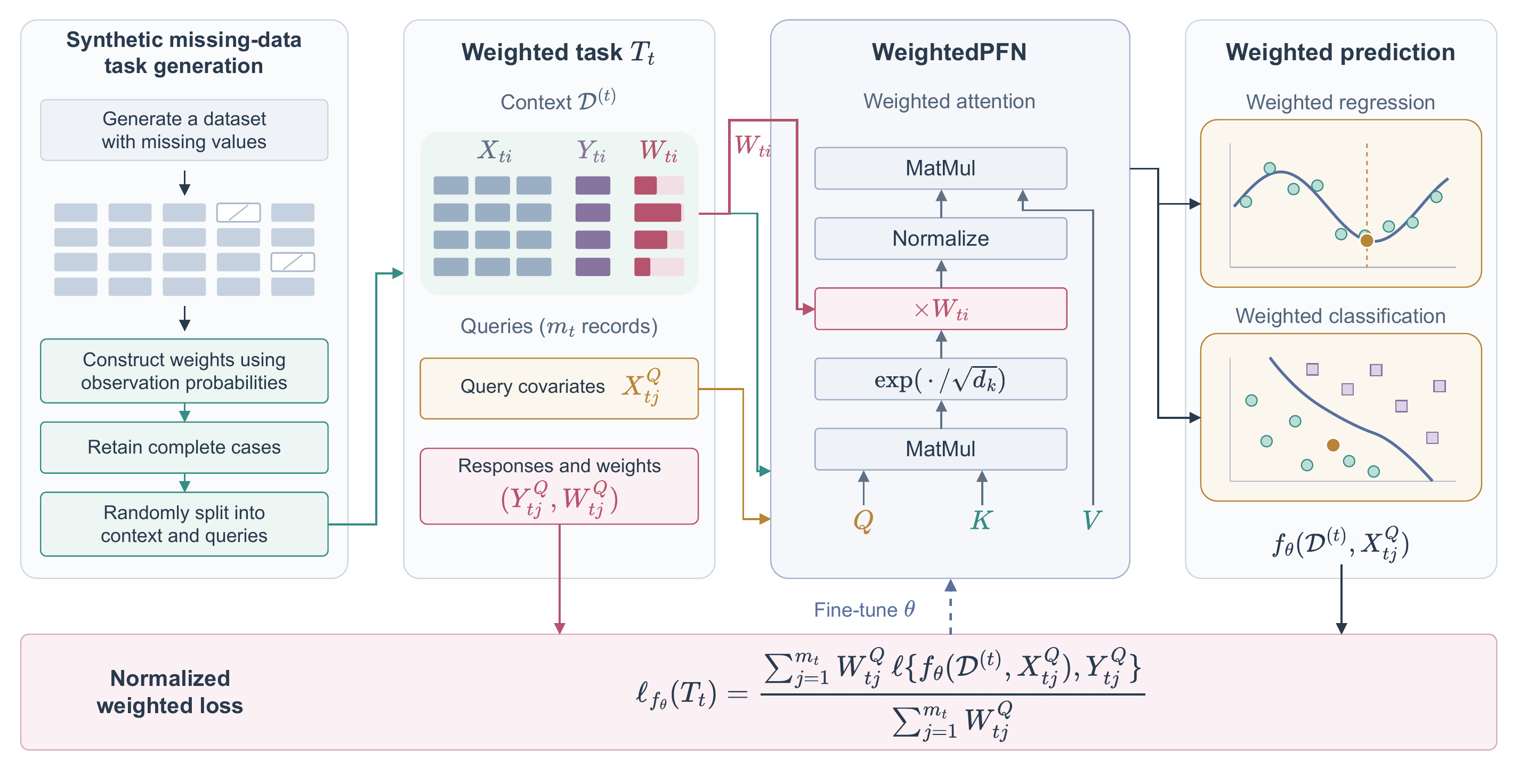}
\caption{Weighted pretraining pipeline. WeightedPFN is fine-tuned on synthetic missing-data tasks with weighted attention and the normalized weighted loss. MatMul denotes matrix multiplication. The operation $\exp(\cdot/\sqrt{d_k})$ divides its input by $\sqrt{d_k}$ and applies the exponential function elementwise. The operation $\times W_{ti}$ multiplies each input entry by its corresponding weight $W_{ti}$.}
\label{fig:weighted-pipeline}
\end{figure}

We pretrain the model exclusively on synthetic missing-data tasks (Example~\ref{ex:intro-missing-data}), which teach it how to use sample weights for regression and classification, and reserve causal tasks (Example~\ref{ex:intro-treatment-selection}) for out-of-distribution evaluation of whether this ability transfers to other weighted problems. Figure~\ref{fig:weighted-pipeline} summarizes the pretraining pipeline. We first describe the construction of the pretraining and evaluation tasks, and then present the model architecture and training objective.

\subsection{Weighted task construction}
\label{sec:data-generation}

\subsubsection{Missing-data prediction tasks for pretraining}
\label{sec:pretraining-tasks}

The missing-data prediction problem in Example~\ref{ex:intro-missing-data} provides the basis for all our synthetic pretraining tasks. It leads to a weighted regression problem when the response is continuous and to a weighted classification problem when the response is categorical, with the same inverse probability weights in both cases. We describe the two formulations in turn.

\paragraph{Weighted regression.} We start with a real-valued response \(Y\), allowing both covariates and responses to be missing. For each observation \(i\), let \(S_{ij}^X=1\) indicate that the \(j\)th covariate is observed, for \(j=1,\ldots,d\), and let \(S_i^Y=1\) indicate that the response is observed. Define \(S_i=S_i^Y\prod_{j=1}^d S_{ij}^X\), so that \(S_i=1\) indicates a complete case. Furthermore, let \(Z_i\) denote the vector of variables that are fully observed across all observations, potentially including the response \(Y_i\). Let $P_o$ denote the original joint distribution of the data before missingness occurs, and let $P_c$ denote the distribution of complete cases, i.e., the distribution conditional on \(S_i=1\), which may differ from $P_o$. Our objective is to estimate the conditional mean under the original distribution, i.e., $\E_{P_o}(Y\mid X=x)$. Because $P_c$ may differ from $P_o$, minimizing the unweighted squared loss over complete cases may fail to recover \(\E_{P_o}(Y\mid X=x)\). Weighting the squared loss corrects this selection bias. Specifically, let $\pi_{S}(z)=\Pp(S=1\mid Z=z)$ denote the complete-case observation probability. Define the weighted samples as
\[(X_i,Y_i,W_i)=\left(X_i,Y_i,\frac{1}{\pi_{S}(Z_i)}\right), \quad S_i=1.\]
Under the positivity condition \(\pi_{S}(Z)>0\) and the missing-at-random assumption \(S\perp (X,Y)\mid Z\) \citep{rubin1976inference}, the weighted conditional mean recovers the conditional mean \(\mathbb{E}_{P_o}(Y\mid X=x)\) of the original data distribution \citep{robins1995semiparametric}.
In real-world applications, the complete-case observation probability $\pi_{S}$ is generally unknown. We can estimate it by treating $S$ as a binary response and the fully observed variables $Z$ as predictors in a probabilistic classification model. The resulting estimated probabilities $\hat\pi_{S}$ are then used to construct the inverse probability weights $W_i=1/\hat\pi_{S}(Z_i)$ for complete cases.

To generate synthetic missing-data regression tasks for pretraining, we simulate complete observations $(X_i,Y_i)$ with real-valued responses and introduce missingness in covariates or responses according to different observation mechanisms $\pi_{S}$. Across tasks, we vary covariate distributions, response functions, outcome noise levels, observation probabilities, and missingness patterns. We also construct weights using the true observation probabilities $\pi_{S}$, randomly perturbed versions of these probabilities $\tilde \pi_{S}$, or their estimates $\hat\pi_{S}$. The complete generation procedure for missing-data regression task types is given in Section~\ref{s:synthetic-missing-data} of the supplementary material.

We consider four missingness mechanisms, named by the variables on which the observation probability $\pi_{S}(Z)$ depends. Under X+Y selection, $Z$ consists of the response and a subset of covariates. Under Y-only selection, $Z$ consists of the response alone. Under X-only selection, $Z$ consists of a subset of covariates alone. Under missing completely at random (MCAR) selection, the observation probability is constant.

\paragraph{Weighted classification.} We now consider the same missing-data setting with a categorical response \(Y\in\{1,\ldots,K\}\). Our objective is to find a classification rule that minimizes the 0--1 loss under the original distribution \(P_o\). The corresponding classification rule is \(\arg\max_{k\in\{1,\ldots,K\}} \Pp_{P_o}(Y=k\mid X=x)\).
Because the complete-case distribution \(P_c\) may differ from \(P_o\), minimizing the unweighted 0--1 loss over complete cases may fail to recover the target rule. As in the regression setting, we reformulate the problem as weighted classification using the observations
\[
(X_i,Y_i,W_i)=\left(X_i,Y_i,\frac{1}{\pi_{S}(Z_i)}\right), \quad S_i=1.
\]
Under the same positivity and missing-at-random conditions as in the regression setting, minimizing the weighted 0--1 loss can recover the classification rule of the original distribution $P_o$. In real-world applications, \(\pi_{S}\) may be estimated as in the regression setting, and each complete case receives the weight \(W_i=1/\hat\pi_{S}(Z_i)\).

To generate synthetic missing-data classification tasks for pretraining, we follow the same data-generating process as for regression, but with categorical responses. The complete generation procedure for missing-data classification task types is given in Section~\ref{s:synthetic-missing-data} of the supplementary material.

\subsubsection{Causal tasks for out-of-distribution evaluation}
\label{sec:causal-tasks}

Beyond missing-data analysis, weighted prediction has many other applications, as discussed in Section~\ref{sec:introduction}. Among these, we use the causal tasks in Example~\ref{ex:intro-treatment-selection} for out-of-distribution evaluation.

\paragraph{Weighted regression.} We begin with CATE estimation through weighted regression. Let \(A\in\mathcal A\) denote the assigned treatment, where \(\mathcal A\) is a finite set of available treatments. For each \(a\in\mathcal A\), let \(R(a)\) denote the potential outcome under treatment \(a\), and let \(R=R(A)\) be the observed outcome. We assume that larger outcomes are preferred. Define the conditional mean potential outcome \(\mu_a(x)=\E\{R(a)\mid X=x\}\) and the propensity score \(\pi_a(x)=\Pp(A=a\mid X=x)\). Under consistency, conditional exchangeability, and positivity, \(\mu_a(x)=\E(R\mid A=a,X=x)\) \citep{hernan2020causal}. These conditions hold in randomized experiments, where the propensity score is also known. For binary treatments, take \(\mathcal A=\{-1,1\}\). The CATE is \(\tau(x)=\mu_1(x)-\mu_{-1}(x),\) whose sign determines which treatment yields the larger conditional mean outcome. The CATE can be estimated through weighted regression without separately estimating the two conditional mean potential outcomes \citep{tian2014simple}. Namely, consider the weighted samples
\[
(X_i,Y_i,W_i)=\left(X_i,\;2A_iR_i,\;\frac{1}{\pi_{A_i}(X_i)}\right),
\]
for which the weighted conditional mean satisfies \({\E(WY\mid X=x)}/{\E(W\mid X=x)}=\tau(x)\). We refer to $2A_iR_i$ as the ordinary modified outcome, and to weighting by $1/\pi_{A_i}(X_i)$ as inverse probability weighting (IPW). We also consider three other weighted sample constructions that target the same CATE. The first replaces the ordinary modified outcome $2A_iR_i$ with the residualized modified outcome \citep{meng2022augmented}. The second replaces IPW with overlap weighting \citep{li2018balancing}. The third makes both replacements. Details are given in Section~\ref{s:causal-regression} of the supplementary material.

\paragraph{Weighted classification.} We next consider learning treatment rules directly through weighted classification. Here, let \(\mathcal A=\{1,\ldots,K\}\) for \(K\) treatments. A treatment rule \(\varphi\) assigns treatment \(\varphi(x)\) to an individual with covariates \(x\), and its value is defined as \(V(\varphi)=\E\{R(\varphi(X))\}\). Our goal is to estimate the optimal treatment rule \(\varphi^*=\arg\max_\varphi V(\varphi)\). This problem can be formulated as weighted classification through outcome-weighted learning (OWL) \citep{zhao2012owl}. Specifically, consider the weighted samples
\[
(X_i,Y_i,W_i)=\left(X_i,\;A_i,\;\frac{R_i+C}{\pi_{A_i}(X_i)}\right),
\]
where \(C\) is any sufficiently large constant such that \(C+R_i\ge 0\) for all \(i\), ensuring all observed weights are nonnegative. The optimal treatment rule can be recovered by minimizing the weighted 0--1 loss: \(\varphi^*=\arg\max_\varphi V(\varphi)=\arg\min_\varphi \E_P\left[W\ind\{\varphi(X)\ne Y\}\right]\) \citep{zhao2012owl}. We also consider other weighted sample constructions that yield the optimal treatment rule using residual weighted learning (RWL) \citep{zhou2017rwl} and doubly robust (DR) learning \citep{zhang2012classification} in Section~\ref{s:rwl-dr} of the supplementary material.

\subsection{Model architecture}
\label{sec:model-architecture}

We incorporate sample weights into the attention mechanism of the Transformer. The resulting weighted attention can replace standard attention in Transformer-based tabular foundation models. In this paper, we apply it to TabPFN-3.5 to obtain WeightedPFN.

To gain intuition for the proposed weighted attention mechanism, we first introduce the notation and describe how a standard attention mechanism operates. Denote the context by $\Dd_n=\{(X_i,Y_i,W_i)\}_{i=1}^n$ and the query covariates by $X^Q$ whose response is unknown. Tabular foundation models with standard attention feed the context without the sample weights, i.e., \(\{(X_i,Y_i)\}_{i=1}^n\), together with the query \(X^Q\) into the Transformer and read the prediction for \(X^Q\) from its output.

To see how a Transformer uses attention to make predictions, consider a single-layer Transformer with one attention head and without embedding, layer normalization, or residual connections. Each context observation is represented by $u_i=(X_i^\top,Y_i)^\top\in\R^{d+1}$, and the query observation by $u^Q=((X^Q)^\top,0)^\top\in\R^{d+1}$, where zero is a placeholder for the unknown query prediction. Linear projections first give the query vectors $q_i^C=B_Qu_i$ and $q^Q=B_Qu^Q$, the key vectors $k_i=B_Ku_i$, and the value vectors $v_i=B_Vu_i$, where $B_Q,B_K,B_V\in\R^{d_k\times(d+1)}$ are matrices learned during pretraining. Stacking all query vectors gives $Q=[q_1^C,\ldots,q_n^C,q^Q]^\top\in\R^{(n+1)\times d_k}$. Similarly, stacking key vectors and value vectors gives $K$ and $V$.

Standard query--key--value (QKV) attention \citep{vaswani2017attention} then computes an output for every row of $Q$. Writing $q_j^\top$ for the $j$th row of $Q$, with $q_{n+1}=q^Q$, we have
\[
\alpha_{ij}
=\frac{\exp\left(k_i^\top q_j/\sqrt{d_k}\right)}
{\sum_{t=1}^n\exp\left(k_t^\top q_j/\sqrt{d_k}\right)},
\qquad
\operatorname{Attention}(Q,K,V)_j=\sum_{i=1}^n\alpha_{ij}v_i,
\]
where $\operatorname{Attention}(Q,K,V)_j$ denotes the attention output vector corresponding to $q_j$ for $j=1,\ldots,n+1$. In the single-layer Transformer, prediction then uses only the final row after these operations, since it corresponds to the query observation. Specifically, the prediction is $g\{\operatorname{Attention}(Q,K,V)_{n+1}\}$, where $g$ is a prediction head typically implemented as a feed-forward network (FFN).

For the final row of the attention output, which corresponds to the query to be predicted, the coefficient \(\alpha_{i,n+1}\) can be interpreted as a measure of similarity between the query \(X^Q\) and the \(i\)th context observation. Using these similarity coefficients to average the value vectors of the context observations then combines context observations for prediction according to their similarity to the query, which is similar to Nadaraya--Watson (NW) kernel regression \citep{nadaraya1964,watson1964}. To incorporate sample weights into attention, we therefore let these weights adjust the similarity coefficients, as in weighted NW regression \citep{wang2010nonparametric}:
% The modified coefficients and the resulting attention mechanism should satisfy four properties: (i) \emph{consistency}: it reduces to standard attention when all sample weights are equal and positive; (ii) \emph{zero-weight exclusion}: observations with zero weights make no contribution to the attention output; (iii) \emph{scale invariance}: multiplying all sample weights by the same positive constant leaves the attention output unchanged; and (iv) \emph{weight-splitting invariance}: replacing an observation with multiple identical copies whose weights sum to the original weight leaves the attention output unchanged.
\[
\alpha_{ij}
=\frac{W_i\exp\left(k_i^\top q_j/\sqrt{d_k}\right)}
{\sum_{t=1}^n W_t\exp\left(k_t^\top q_j/\sqrt{d_k}\right)},
\qquad
\operatorname{WeightedAttention}(Q,K,V;W)_j=\sum_{i=1}^n\alpha_{ij}v_i,
\]
where $\operatorname{WeightedAttention}(Q,K,V;W)_j$ denotes the weighted attention output vector and $W_i\geq 0$ is the sample weight associated with the $i$th context observation. 

The single-layer, single-head Transformer above only serves to convey the intuition behind weighted attention. In a full tabular foundation model with multiple layers and attention heads, such as TabPFN-3.5, the base model of our WeightedPFN, we apply the same modification to every attention head in every layer.

% This weighted attention mechanism satisfies the four properties described above while preserving the standard QKV structure.
% We also incorporate sample weights into covariate preprocessing, as detailed in Section~\ref{sec:implementation}.

\subsection{Training objective}
\label{sec:training-objective}

Having specified how the model uses sample weights, we now describe how its parameters are learned. We fine-tune the model on synthetic missing-data tasks $T_1,\ldots,T_M$ generated as in Section~\ref{sec:pretraining-tasks}. For task $T_t$, we randomly split the weighted complete cases into a context $\Dd^{(t)}=(X_{ti},Y_{ti},W_{ti})_{i=1}^{n_t}$ and $m_t$ query samples $(X_{tj}^Q,Y_{tj}^Q,W_{tj}^Q)_{j=1}^{m_t}$, where the context size $n_t$ and query size $m_t$ vary across tasks. Given $\Dd^{(t)}$, the model with parameters $\theta$ produces a prediction $f_\theta(\Dd^{(t)},X_{tj}^Q)$ for each query covariate $X_{tj}^Q$. For weighted regression, this is a scalar prediction. For weighted classification, it is a vector of predicted class probabilities $p_\theta(\Dd^{(t)},X_{tj}^Q)=(p_{\theta,1}(\Dd^{(t)},X_{tj}^Q),\ldots,p_{\theta,K}(\Dd^{(t)},X_{tj}^Q))$, and the corresponding weighted classification rule is $\arg\max_k p_{\theta,k}(\Dd^{(t)},X_{tj}^Q)$.

The training objective is an empirical version of the weighted population risk. For a predictor $f$, the normalized weighted loss on task $T_t$ is the weighted average of the losses on its queries,
\begin{equation}\label{eq:risk}
 \ell_f(T_t)
 =\frac{\sum_{j=1}^{m_t} W_{tj}^Q\,\ell\{f(\Dd^{(t)},X_{tj}^Q),Y_{tj}^Q\}}
     {\sum_{j=1}^{m_t} W_{tj}^Q},
\end{equation}
and the overall loss is its average $M^{-1}\sum_{t=1}^M\ell_{f_\theta}(T_t)$ over the $M$ tasks. Normalizing by the total query weight \(\sum_{j=1}^{m_t} W_{tj}^Q\) makes tasks with different weight scales contribute comparably.

The loss $\ell$ depends on the type of task. For weighted regression, $\ell(a,b)=(a-b)^2$ is the squared-error loss. For weighted classification, the 0--1 loss is non-differentiable and difficult to optimize, so we use cross-entropy as a differentiable surrogate. For an observation with class label $y$, the cross-entropy loss is $\ell(p,y)=-\log p_y$, where $p=(p_1,\ldots,p_K)$ is the vector of class probabilities and $p_y$ is the probability assigned to class $y$. Cross-entropy loss is Fisher consistent for the 0--1 loss \citep{zhang2004multicategory}. Specifically, the weighted cross-entropy risk $-\E_P[W\log p_Y(X)]$ is minimized by $p_k^*(x)=\E_P[W\ind(Y=k)\mid X=x]/\E_P[W\mid X=x]$ for $k=1,\ldots,K$. Thus, choosing $\arg\max_k p_k^*(x)$ recovers the desired weighted classification rule.

\section{Theoretical properties}
\label{sec:theory}

In this section, we study how accurately a pretrained weighted in-context learner predicts on new tasks. We focus on weighted regression with squared-error loss, for which the target of a task with distribution $P$ is the weighted conditional mean $m_P$. 
% We analyze a single-head softmax Transformer that uses the weighted attention of Section~\ref{sec:model-architecture} and is pretrained by minimizing the training objective of Section~\ref{sec:training-objective} with the squared-error loss. 
% The tasks met at evaluation may follow a different distribution from the pretraining tasks. We bound the excess weighted risk of the pretrained predictor on evaluation tasks, relative to the regression function of each task, and give conditions under which this risk attains the minimax rate of nonparametric regression based on the $n$ context records.

\label{sec:model}\label{sec:risk}
We consider contexts $\Dd_n$ and query covariates $X$ of the form
\begin{equation}\label{eq:data}
 \Dd_n=(X_i,Y_i,W_i)_{i=1}^n,\qquad
 X_i,X\in[0,1]^d,\quad |Y_i|\le B_Y,\quad
 0<W_i\le B_W ,
\end{equation}
where $B_Y,B_W\ge1$ are fixed constants, and the responses and weights of query samples satisfy the same bounds.

We study the class $\Ff$ of single-head Transformers $f_\theta(\Dd_n,X)$ with $L$ blocks. Affine embeddings first map each context pair $(X_i,Y_i)$ and the query covariates $X$ to $d_k$-dimensional vectors, called the initial token representations. Within each block, linear projections of the token representations produce the query, key, and value vectors introduced in Section~\ref{sec:model-architecture}, which we also take to have dimension $d_k$. The block then applies the weighted attention mechanism in Section~\ref{sec:model-architecture}, followed by an FFN with $D_{\mathrm{FFN}}$ hidden layers of width at most $W_{\mathrm{FFN}}$ and GELU activations \citep{hendrycks2016}. The prediction is clipped to $[-B_Y,B_Y]$. Section~\ref{sec:network} of the supplementary material gives the precise architecture we analyze theoretically. Its core components, including the weighted attention and the GELU activation, are the same as in the WeightedPFN used in our experiments. To keep the analysis tractable, as in previous theoretical studies of in-context learning \citep{garg2022what,ching2026}, we omit some components, including multiple heads and layer normalization.

A task law $P$ is a distribution of $(X,Y,W)$ satisfying the bounds in \eqref{eq:data}. Let $\mathcal P$ be a family of task laws, and let $\Pi^{\mathrm{train}}$ and $\Pi^{\mathrm{eval}}$ be the pretraining and evaluation distributions on $\mathcal P$. They may assign different probabilities to regression functions, designs, and weight mechanisms. For the theoretical analysis, we fix the context and query sizes at $n$ and $m\ge1$, respectively. A complete task is generated by first drawing $P$ from the corresponding task distribution, then drawing $n$ context samples and $m$ query samples independently from $P$. The pretraining tasks $T_1,\ldots,T_M$ consist of a context and queries as in Section~\ref{sec:training-objective}, with $n_t=n$ and $m_t=m$. 
% This hierarchical description is consistent with task-level analyses of in-context learning \citep{bai2023}.

\vspace{0.3ex}
\textbf{Assumption 1:} The $M$ pretraining tasks are generated independently with task laws $P_1,\ldots,P_M$ drawn from $\Pi^{\mathrm{train}}$. Independently of pretraining, an evaluation task law $P$ is drawn from $\Pi^{\mathrm{eval}}$, and its context and queries are generated as above.
\vspace{0.3ex}

% For a task law $P$, we set $m_P(x)=0$ where $\E_P(W\mid X=x)=0$. 
Write $\widehat\Risk_{n,M}^{\mathrm{train}}(f)=M^{-1}\sum_{t=1}^M\ell_f(T_t)$ for the pretraining loss, with $\ell_f$ in \eqref{eq:risk} and the squared-error loss. Let $\widehat\theta_M$ denote the parameters returned by an optimization algorithm applied to the pretraining loss $\widehat\Risk_{n,M}^{\mathrm{train}}(f_\theta)$. Write $\widehat f_M=f_{\widehat\theta_M}\in\Ff$ and $\Eopt=\widehat\Risk_{n,M}^{\mathrm{train}}(\widehat f_M)-\inf_{f\in\Ff}\widehat\Risk_{n,M}^{\mathrm{train}}(f)$, where the optimization error $\Eopt$ depends on the optimization algorithm. 

Let $\E_{\mathrm{eval}}$ denote expectation over a task law $P\sim\Pi^{\mathrm{eval}}$, its context, and an independent query. Define the evaluation risk and its oracle benchmark as
\begin{equation}\label{eq:population-risk}
 \Risk_n^{\mathrm{eval}}(f)=\E_{\mathrm{eval}}\Bigl[\frac{W}{\E_P W}\{f(\Dd_n,X)-Y\}^2\Bigr],\qquad
 \Risk_*^{\mathrm{eval}}=\E_{\mathrm{eval}}\Bigl[\frac{W}{\E_P W}\{Y-m_P(X)\}^2\Bigr].
\end{equation}

Our prediction criterion is the excess risk $\Risk_n^{\mathrm{eval}}(\widehat f_M)-\Risk_*^{\mathrm{eval}}$, which equals the average over $P\sim\Pi^{\mathrm{eval}}$ of $\E_P[W\{\widehat f_M(\Dd_n,X)-m_P(X)\}^2]/\E_PW$. This criterion is natural for two reasons. First, because the weighted conditional mean $m_P$ is the target of each task, the excess risk measures how far the prediction is from this target. Second, normalizing by the mean weight $\E_PW$ makes the tasks whose weights are on different scales contribute comparably to the average over evaluation tasks.

\label{sec:assumptions}
We now state the assumptions. For a multi-index $\nu=(\nu_1,\ldots,\nu_d)\in\mathbb N_0^d$ with $|\nu|=\sum_j\nu_j$, we write $\partial^\nu$ as shorthand for the mixed partial derivative $\partial^{|\nu|}/(\partial x_1^{\nu_1}\cdots\partial x_d^{\nu_d})$.

\vspace{0.3ex}
\textbf{Assumption 2:} For a fixed $B_p\ge1$ and every task law \(P\), the covariate vector \(X\) has a density $p_P$ on $[0,1]^d$ with $B_p^{-1}\le p_P(x)\le B_p$ for all $x\in[0,1]^d$.

\textbf{Assumption 3:} For a fixed $B_a\ge1$, the conditional mean weight \(\E_P[W\mid X=x]\) for every task law \(P\) satisfies $\E_P[W\mid X=x]\ge B_a^{-1}$ for all $x\in[0,1]^d$.

\textbf{Assumption 4:} For fixed $s>0$ and $B_m\ge1$, the weighted regression function $m_P$ for every task law \(P\) lies in the H\"older ball $\Hh^s([0,1]^d;B_m)$. That is, with $k=\lceil s\rceil-1$, the partial derivatives $\partial^\nu m_P$ with $|\nu|\le k$ extend continuously to $[0,1]^d$ and satisfy
\[
 \max_{|\nu|\le k}\sup_{x\in[0,1]^d}|\partial^\nu m_P(x)|\le B_m,
 \qquad
 \max_{|\nu|=k}\sup_{\substack{x,z\in[0,1]^d\\x\ne z}}
 \frac{|\partial^\nu m_P(x)-\partial^\nu m_P(z)|}
      {\norm{x-z}_2^{s-k}}\le B_m.
\]
\vspace{0.3ex}

\textbf{Assumption 5:} For a fixed $\kappa\ge1$, the evaluation task distribution is absolutely continuous with respect to the pretraining task distribution, with Radon--Nikodym derivative bounded by $\kappa$:
\begin{equation}\label{eq:task-coverage}
 \Pi^{\mathrm{eval}}\ll\Pi^{\mathrm{train}},\qquad
 \frac{d\Pi^{\mathrm{eval}}}{d\Pi^{\mathrm{train}}}\le\kappa .
\end{equation}
\vspace{0.3ex}

Assumptions~2--4 are standard conditions for establishing convergence rates in nonparametric regression \citep{stone1982}. 
% Assumption~2 ensures that context records fall every covariate value. Assumption~3 ensures that records carry non-negligible weight, so that the weighted target $m_P$ is determined by the data at every covariate value; for example, the inverse probability weights $W=1/\pi_{S}(X,Y)$ of Section~\ref{sec:pretraining-tasks} satisfy $W\ge1$, so that Assumption~3 holds with $B_a=1$. Assumption~4 states that $m_P$ has smoothness $s$; smoother regression functions can be estimated more accurately from the same number of records. 
Assumption~5 requires that pretraining covers the kinds of tasks encountered at evaluation. This assumption motivates generating diverse synthetic pretraining tasks. 
% The causal applications below test transfer empirically. Coverage of their task laws by the missing-data prior has not been established.

\label{sec:main-result}
We now provide the main theorem:

\begin{theorem}\label{thm:combined}
Suppose Assumptions~1--5 hold, and let $r_s=\max\{1,\lceil s\rceil-1\}$.
There are constants $C_1,C_2,C_3\ge1$, depending only on $d$, and
$C_s,C_4,C_5\ge1$, depending only on $d$, $s$,
$B_Y,B_W,B_p,B_a$, and $B_m$, such that the following holds: if $d_k\ge\lceil C_1r_s^d\rceil$, $L\ge C_2\{1+\log(r_s+1)\}$, and
$W_{\mathrm{FFN}}\ge C_3r_s^{2d}$, then, for any $\delta\in(0,1)$, with probability at least
$1-\delta$ over the $M$ pretraining tasks, the predictor
$\widehat f_M$ satisfies
\begin{equation}\label{eq:combined}
 \begin{split}
 &\Risk_n^{\mathrm{eval}}(\widehat f_M)-\Risk_*^{\mathrm{eval}}
 \le\kappa C_sn^{-2s/(2s+d)}+\kappa C_se^{-L/C_s}\\
 &\qquad+2\kappa C_4\left\{
 L^2D_{\mathrm{FFN}}^2(n+1)^2(W_{\mathrm{FFN}}+d_k)^4
       \sqrt{\frac{\log(2M)}{M}}
       +\sqrt{\frac{\log(2/\delta)}{M}}\right\}\\
 &\qquad+\frac{2\kappa C_5}m+\kappa\Eopt.
 \end{split}
\end{equation}
\end{theorem}

Theorem~\ref{thm:combined} shows that a pretrained Transformer can learn a new weighted regression task in context. After pretraining, the same Transformer is applied to the context of each evaluation task without any further training or parameter update, and its excess risk on evaluation tasks is controlled by \eqref{eq:combined}. We now explain what each term represents.

% The proof is given in Section~\ref{s:combined} of the supplementary material, where $C_4$ and $C_5$ are the constants of Theorems~\ref{thm:pretrain} and~\ref{thm:query}. The bound \eqref{eq:combined} separates the excess risk into errors that arise within a single task, errors that arise from pretraining, and the effect of task distribution shift. We explain each term in turn.

The first two terms describe what happens within a single task. The first term, $n^{-2s/(2s+d)}$, is the statistical error of learning a task from its $n$ context samples. For a new task, our method sees only the $n$ context samples, and no estimator built from them can achieve a faster rate uniformly over regression functions with smoothness $s$ \citep{stone1982}. Thus, learning in context loses nothing in convergence rate relative to the best estimator fitted directly to the task.
% Weighting does not change this rate. By the definition of $m_P$, the weighted residual $W\{Y-m_P(X)\}$ has mean zero at every covariate value, so weighting introduces no bias, and the weights affect only the constant $C_s$.
The second term, $e^{-L/C_s}$, is the computation error. Our Transformer-based method does not fit the new task by updating its parameters. Instead, its forward pass can be viewed as carrying out the fitting, and each of the $L$ blocks refines the prediction in the manner of one step of an iterative algorithm \citep{vonoswald2023,bai2023,ching2026}. A deeper Transformer performs more such steps, so this error decreases as $L$ increases.

% The next three terms concern pretraining, which must produce parameters that encode a good fitting procedure for all tasks.
 The third term is the pretraining error, which arises because the parameters of our method are learned from $M$ tasks generated from the pretraining distribution rather than from the whole pretraining distribution. As in a uniform law of large numbers, it decreases at a rate of $\{\log(M)/M\}^{1/2}$. The fourth term, $2\kappa C_5/m$, arises because the training loss \eqref{eq:risk} is normalized using the empirical mean weight \(\sum_{j=1}^{m}W_j^{Q}/m\) rather than the population mean weight $\E_PW$. The last term, $\kappa\Eopt$, is the optimization error of pretraining, which depends on the optimizer rather than on the statistical problem.

Task distribution shift enters only through the factor $\kappa$. In-context learning transfers what is learned on pretraining tasks to new tasks. Because Assumption~5 ensures that the new tasks are covered by the pretraining distribution, a Transformer that performs well on average over pretraining tasks loses at most a factor $\kappa$ on evaluation tasks. When $\Pi^{\mathrm{eval}}=\Pi^{\mathrm{train}}$, the bound holds with $\kappa=1$.

The next corollary chooses the network size and the numbers of queries and pretraining tasks so that our method can achieve the minimax rate.

\begin{corollary}[Minimax rate]\label{cor:minimax}
Suppose Assumptions~1--5 hold, and let $r_s$, $C_1$, and $C_3$ be as
in Theorem~\ref{thm:combined}. Take $D_{\mathrm{FFN}}=1$,
$d_k=\lceil C_1r_s^d\rceil$, $L=\lceil C_L\log(n+1)\rceil$, and $W_{\mathrm{FFN}}=\lceil C_3r_s^{2d}\rceil$, where $C_L\ge1$ is a constant
depending only on $d$, $s$, $B_Y,B_W,B_p,B_a$, and $B_m$, given in
Section~\ref{s:minimax} of the supplementary material. If, as $n\to\infty$,
$m\ge n^{2s/(2s+d)}$,
$M/\log(2M)\ge(n+1)^{4+4s/(2s+d)}\log^4(n+1)$, and
$\Eopt=O_p(n^{-2s/(2s+d)})$, then
\begin{equation}\label{eq:minimax}
 \Risk_n^{\mathrm{eval}}(\widehat f_M)-\Risk_*^{\mathrm{eval}}
 =O_p\bigl(n^{-2s/(2s+d)}\bigr).
\end{equation}
\end{corollary}

% The proof is given in Section~\ref{s:minimax} of the supplementary material.
 Corollary~\ref{cor:minimax} shows that our method attains the minimax rate for weighted regression on new tasks without retraining. This is the optimal convergence rate achievable from $n$ context observations. 
%  We now discuss whether its conditions are easy to satisfy.

The conditions on the Transformer architecture are mild. The token dimension $d_k$ and the FFN width $W_{\mathrm{FFN}}$ do not grow with $n$ and depend only on $d$ and $s$, and one hidden layer per FFN, $D_{\mathrm{FFN}}=1$, suffices. The number of attention blocks $L$ grows only logarithmically in $n$.
% The required token dimension and width can nevertheless be large when $d$ or $s$ is large, because they grow like $r_s^d$ and $r_s^{2d}$.
The condition on the number of queries per task during pretraining, $m$, is also mild. Since $2s/(2s+d)<1$, it requires fewer queries than context samples $n$. For example, splitting each pretraining task into context and queries in a fixed proportion, as in our implementation, satisfies it for all sufficiently large $n$. The condition on the number of pretraining tasks is the most demanding: $M$ must grow polynomially in $n$, roughly $n^{4+4s/(2s+d)}$ up to logarithmic factors. Because pretraining tasks are generated synthetically, this condition can be met by generating more tasks. The condition on the optimization error \(\Eopt\) is determined by the optimization algorithm.

\section{Experiments}
\label{sec:experiments}

In this section, we first describe the implementation details for pretraining our weighted regression and classification models on synthetic missing-data tasks. We then introduce the competing methods and compare their performance with our models on two groups of tasks based on real datasets. The first consists of missing-data tasks, while the second consists of causal tasks used to assess out-of-distribution task generalization.

\subsection{Implementation}
\label{sec:implementation}

We generate synthetic tasks with 128 to 1,024 observations and 4 to 16 covariates before imposing missingness. The task generator samples the X+Y, Y-only, X-only, and MCAR mechanisms of Section~\ref{sec:pretraining-tasks} with probabilities 0.4, 0.4, 0.1, and 0.1, respectively. Weight construction uses oracle, perturbed, or TabPFN-estimated observation probabilities, as detailed in Section~\ref{s:synthetic-missing-weights} of the supplementary material. Each task supplies 70\% of its weighted samples as context and the remainder as queries. We initialize from the released parameters of TabPFN-3.5 and use AdamW \citep{loshchilov2019decoupled} with a learning rate of $10^{-5}$, weight decay of $0.01$, gradient-norm clipping at 1, and a batch of 16 tasks per update. We use the checkpoint after 800 updates. Thus, WeightedPFN is fine-tuned on 12,800 synthetic datasets. Fine-tuning takes about 0.5 hours on a single NVIDIA H100 GPU for both weighted regression and weighted classification.

\subsection{Comparison methods}
\label{sec:baselines}

\emph{Competing methods for weighted regression.} We compare our WeightedPFN regressor with the unweighted TabPFN-3.5 regressor and five baselines: weighted ridge linear regression (Ridge), weighted support vector regression with a radial basis function kernel (SVR; \citealp{smola2004tutorial}), weighted random forests (RF; \citealp{breiman2001random}), weighted LightGBM (LightGBM; \citealp{ke2017lightgbm}), and a weighted multilayer perceptron (MLP; \citealp{rumelhart1986learning}). Each of the five baselines uses $W_i$ as a sample weight in its training loss. Hyperparameters are chosen by five-fold cross-validation within the context, minimizing weighted held-out prediction error. The tuned hyperparameters for these methods are listed in Section~\ref{s:regression-tuning} of the supplementary material.

\emph{Competing methods for weighted classification.} We compare our WeightedPFN classifier with the unweighted TabPFN-3.5 classifier and five baselines: weighted logistic regression (Logistic), a weighted support vector machine with a radial basis function kernel (SVM; \citealp{cortes1995support}), weighted random forests (RF), weighted LightGBM (LightGBM), and a weighted multilayer perceptron (MLP). Each of the five baselines uses $W_i$ as a sample weight in its training loss. Hyperparameters are chosen by five-fold cross-validation within the context, maximizing weighted held-out accuracy. The tuned hyperparameters for these methods are listed in Section~\ref{s:classification-tuning} of the supplementary material.

For both regression and classification, TabPFN-3.5 receives the same context covariates and responses as WeightedPFN but does not use the sample weights. Both in-context models make predictions without updating their parameters on the evaluation tasks.

\subsection{Missing-data prediction}
\label{sec:exp-talent}

We use 110 regression datasets and 156 classification datasets from the TALENT benchmark collection to evaluate competing methods in missing-data prediction tasks \citep{liu2025talent}. For each dataset, we employ five-fold cross-validation to evaluate the performance of different methods. Specifically, the dataset is divided into five folds, with four folds used for training and the remaining fold held out for evaluation. Within each training set, we introduce missingness under the four mechanisms used in synthetic pretraining: X+Y, Y-only, X-only, and MCAR. The sampling functions and masking procedure are detailed in Section~\ref{s:real-missing-data} of the supplementary material. We then construct weighted regression and classification samples from the complete cases, with weights $W_i=1/\hat\pi_{S}(Z_i)$. Specifically, $\hat\pi_{S}$ is estimated through five-fold cross-fitting using TabPFN-3.5. The resulting weights are therefore estimated rather than known, reflecting the setting encountered in real-world missing-data applications. The validation fold remains complete, and the performance of each method is averaged across the five folds.

% {For regression, we use up to 10,000 complete observations per dataset. For both regression and classification, TabPFN estimates observation probabilities by five-fold cross-fitting within each outer training set under all four mechanisms.}

For regression, our objective is to estimate the conditional mean under the original data distribution $P_o$. Since the validation folds are kept complete, we evaluate prediction error using unweighted root mean squared error (RMSE), scaled by the response standard deviation (SD) to allow comparisons across datasets. Specifically, let $\mathcal I_b$ denote the validation set in fold $b$, and let $\hat f^{(-b)}$ denote the predictor learned from the other four folds. We define the five-fold cross-validated scaled RMSE as
\[
\frac{\widehat{\operatorname{RMSE}}(\hat f)}{\hat\sigma_Y}
=\frac{1}{5\hat\sigma_Y}\sum_{b=1}^5
\left[\frac{1}{|\mathcal I_b|}\sum_{i\in\mathcal I_b}
\{\hat f^{(-b)}(X_i)-Y_i\}^2\right]^{1/2},
\]
where $\hat\sigma_Y$ is the standard deviation of the responses. Lower scaled RMSE indicates better predictive performance.

For classification, our objective is to learn a classification rule that minimizes the 0--1 loss under the original data distribution $P_o$. Since the validation folds are kept complete, we evaluate each method using unweighted classification accuracy. With $\hat f^{(-b)}$ denoting the classification rule learned from the other four folds, we define the five-fold cross-validated accuracy as
\[
\widehat{\operatorname{Acc}}(\hat f)
=\frac{1}{5}\sum_{b=1}^5\frac{1}{|\mathcal I_b|}
\sum_{i\in\mathcal I_b}\ind\{\hat f^{(-b)}(X_i)=Y_i\}.
\]

\begin{figure}[tbp]
\centering
\includegraphics[width=\linewidth]{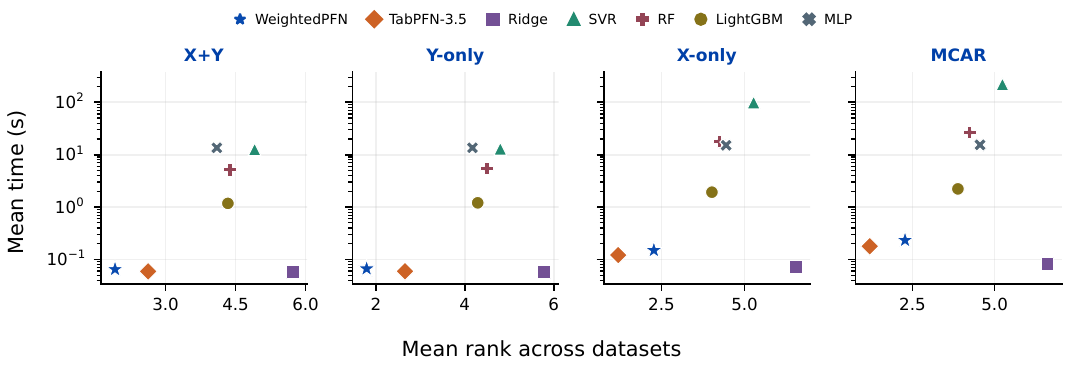}
\caption{{Mean RMSE rank and runtime (log scale) on 110 TALENT regression datasets. Lower values are better.}}
\label{fig:missing_reg}
\end{figure}

\begin{figure}[tbp]
\centering
\includegraphics[width=\linewidth]{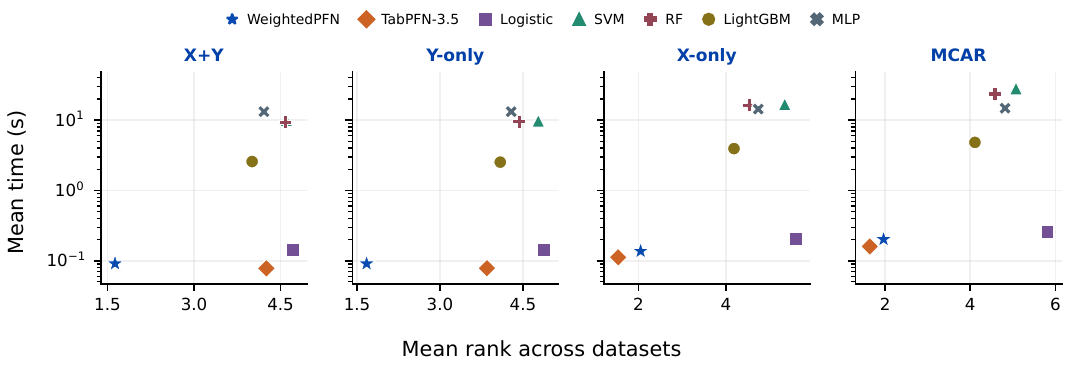}
\caption{{Mean accuracy rank and runtime (log scale) on 156 TALENT classification datasets. Lower values are better.}}
\label{fig:missing_clf}
\end{figure}

Figures~\ref{fig:missing_reg} and~\ref{fig:missing_clf} show the mean rank and mean runtime across datasets, and Table~\ref{tab:talent-performance} reports scaled RMSE for regression and accuracy as a percentage for classification under each missingness mechanism. 

\begin{table}[!htbp]
\centering\singlespacing
\caption{{TALENT performance (mean $\pm$ SD across datasets). Lower scaled RMSE and higher accuracy are better. Bold indicates the best mean in each column.}}\label{tab:talent-performance}
\fontsize{9}{11}\selectfont\setlength{\tabcolsep}{1.35pt}
\begin{tabular*}{\linewidth}{@{\extracolsep{\fill}}lcccc@{}}
\toprule
\multicolumn{5}{@{}l}{\textbf{TALENT regression}}\\
 & \multicolumn{4}{c}{Scaled RMSE (mean $\pm$ SD)}\\
\cmidrule(lr){2-5}
Method & X+Y & Y-only & X-only & MCAR\\
\midrule
WeightedPFN & $\mathbf{0.547}\,{\pm}\,0.310$ & $\mathbf{0.554}\,{\pm}\,0.316$ & $0.474\,{\pm}\,0.298$ & $0.458\,{\pm}\,0.295$\\
\addlinespace[1.5pt]
TabPFN-3.5 & $0.583\,{\pm}\,0.378$ & $0.585\,{\pm}\,0.367$ & $\mathbf{0.467}\,{\pm}\,0.294$ & $\mathbf{0.448}\,{\pm}\,0.289$\\
\addlinespace[1.5pt]
Ridge & $0.728\,{\pm}\,0.249$ & $0.767\,{\pm}\,0.319$ & $0.704\,{\pm}\,0.247$ & $0.692\,{\pm}\,0.243$\\
SVR & $0.652\,{\pm}\,0.274$ & $0.667\,{\pm}\,0.300$ & $0.584\,{\pm}\,0.283$ & $0.562\,{\pm}\,0.280$\\
RF & $0.615\,{\pm}\,0.282$ & $0.634\,{\pm}\,0.306$ & $0.534\,{\pm}\,0.283$ & $0.513\,{\pm}\,0.279$\\
LightGBM & $0.612\,{\pm}\,0.292$ & $0.629\,{\pm}\,0.312$ & $0.528\,{\pm}\,0.292$ & $0.509\,{\pm}\,0.290$\\
MLP & $0.629\,{\pm}\,0.287$ & $0.678\,{\pm}\,0.460$ & $0.555\,{\pm}\,0.291$ & $0.540\,{\pm}\,0.292$\\
\bottomrule
\end{tabular*}\par\medskip
\begin{tabular*}{\linewidth}{@{\extracolsep{\fill}}lcccc@{}}
\toprule
\multicolumn{5}{@{}l}{\textbf{TALENT classification}}\\
 & \multicolumn{4}{c}{Accuracy (\%) (mean $\pm$ SD)}\\
\cmidrule(lr){2-5}
Method & X+Y & Y-only & X-only & MCAR\\
\midrule
WeightedPFN & $\mathbf{81.28}\,{\pm}\,14.93$ & $\mathbf{79.33}\,{\pm}\,16.03$ & $83.52\,{\pm}\,13.81$ & $84.18\,{\pm}\,13.74$\\
\addlinespace[1.5pt]
TabPFN-3.5 & $75.74\,{\pm}\,18.48$ & $75.84\,{\pm}\,18.04$ & $\mathbf{83.68}\,{\pm}\,13.71$ & $\mathbf{84.27}\,{\pm}\,13.71$\\
\addlinespace[1.5pt]
Logistic & $76.25\,{\pm}\,16.33$ & $73.95\,{\pm}\,17.58$ & $77.58\,{\pm}\,16.10$ & $77.73\,{\pm}\,16.18$\\
SVM & $77.29\,{\pm}\,15.66$ & $75.17\,{\pm}\,16.77$ & $79.25\,{\pm}\,15.14$ & $79.79\,{\pm}\,15.07$\\
RF & $77.31\,{\pm}\,16.40$ & $75.40\,{\pm}\,17.15$ & $80.14\,{\pm}\,15.47$ & $80.91\,{\pm}\,15.17$\\
LightGBM & $77.85\,{\pm}\,16.25$ & $75.86\,{\pm}\,17.18$ & $80.53\,{\pm}\,15.30$ & $81.36\,{\pm}\,15.05$\\
MLP & $77.42\,{\pm}\,16.02$ & $75.28\,{\pm}\,17.00$ & $79.58\,{\pm}\,15.51$ & $80.08\,{\pm}\,15.43$\\
\bottomrule
\end{tabular*}
\end{table}

Figures~\ref{fig:missing_reg}--\ref{fig:missing_clf} and Table~\ref{tab:talent-performance} show that WeightedPFN achieves the best mean rank and the best prediction performance for both regression and classification under X+Y and Y-only selection. Under X-only and MCAR selection, TabPFN-3.5 performs best, with WeightedPFN close behind. This is expected, since under X-only and MCAR selection, the conditional distribution of $Y$ given $X$ is preserved, so weighting does not change the optimal predictor. The results suggest a trade-off: WeightedPFN gains substantially under X+Y and Y-only selection at the cost of a slight loss under X-only and MCAR selection. For classification in Table~\ref{tab:talent-performance}, for example, mean accuracy increases by 5.54 and 3.49 percentage points under X+Y and Y-only selection and decreases by at most 0.16 percentage points under X-only and MCAR selection. We consider this trade-off acceptable because the loss is small. In terms of computational cost, WeightedPFN has runtimes comparable to those of the linear baselines and substantially lower than those of the nonlinear baselines. Its favorable performance and low runtime place it near the lower-left corner of Figures~\ref{fig:missing_reg} and~\ref{fig:missing_clf}. 

% {On TALENT regression, the average ranks across the four mechanisms are 2.06 for WeightedPFN and 1.92 for unweighted TabPFN-3.5. Under X+Y and Y-only selection, the mean scaled RMSEs are 0.547 and 0.554 for WeightedPFN and 0.583 and 0.585 for the unweighted model (Figure~\ref{fig:missing_reg}). On TALENT classification, the average ranks across the four mechanisms are 1.83 for WeightedPFN and 2.82 for unweighted TabPFN-3.5 (Figure~\ref{fig:missing_clf}). Under X+Y and Y-only selection, WeightedPFN attains mean accuracies of 81.28\% and 79.33\%, compared with 75.74\% and 75.84\% for the unweighted model.}

\subsection{Treatment selection}
\label{sec:exp-causal}

{We use 78 publicly available datasets from randomized experiments to assess out-of-distribution generalization of our WeightedPFN to causal tasks. Most of them are randomized clinical trials from RCT\_Bench \citep{shao2026covariates}. The datasets used in our evaluation are listed in Section~\ref{s:causal-datasets} of the supplementary material. Of the 78 datasets, 55 have two arms and 23 have three or more arms.}

{To evaluate weighted regression tasks, we form binary treatment pairs and apply covariate-dependent sampling within each pair to induce covariate-dependent treatment propensities. Without this step, balanced randomized experiments have constant propensity scores of $0.5$, so the CATE estimation task in Section~\ref{sec:causal-tasks} reduces to unweighted regression and does not assess the model's ability to use sample weights. We retain 148 treatment-pair tasks with at least 25 selected observations per arm. For each task, we construct the four weighted regression samples described in Section~\ref{sec:causal-tasks}, which combine ordinary or residualized modified outcomes with IPW or overlap weighting. All these constructions target the same CATE.}

{To evaluate weighted classification, we use the observed treatment assignments without additional covariate-dependent sampling. The sample weights depend on outcome information and can therefore vary even when treatment propensities are constant. We construct weighted classification samples using DR and RWL for 163 binary treatment-pair tasks. For OWL, we use all available arms jointly. Because multiple outcomes from one dataset define separate tasks, this yields 84 tasks from the 78 datasets. All three constructions target the optimal treatment rule. Further details on DR and RWL are provided in Section~\ref{s:rwl-dr} of the supplementary material.}

For regression, our objective is to estimate the CATE. Since the true CATE is unknown in real datasets, we cannot directly evaluate its estimation error. We therefore construct a treatment rule from the sign of the estimated CATE and evaluate its policy value. For classification, the model directly learns a treatment rule, which we evaluate using the same criterion. In both cases, we estimate policy value on held-out evaluation observations using a doubly robust estimator \citep{zhang2012robust}. Specifically, let $\mathcal I_b$ denote the validation set in fold $b$, and let $\hat\varphi^{(-b)}$ denote the treatment rule learned from the other four folds. We define the five-fold cross-validated scaled policy value as
\[
\frac{\widehat V(\hat\varphi)}{\hat\sigma_R}
=\frac{1}{5\hat\sigma_R}\sum_{b=1}^5
\frac{1}{|\mathcal I_b|}\sum_{i\in\mathcal I_b}
\left[
\hat\mu_{\hat\varphi^{(-b)}(X_i)}^{(-b)}(X_i)
+\frac{\ind\{A_i=\hat\varphi^{(-b)}(X_i)\}}{\pi_{A_i}(X_i)}
\left\{R_i-\hat\mu_{A_i}^{(-b)}(X_i)\right\}
\right],
\]
where $\hat\sigma_R$ is the standard deviation of the observed outcomes, $\hat\mu_a^{(-b)}$ is the outcome model fitted on the training folds, and $\pi_a$ is the known propensity score. Outcome models and propensity estimation are described in Section~\ref{s:causal-outcome-estimation} of the supplementary material. Higher values indicate better treatment selection.

\begin{figure}[tbp]
\centering
\includegraphics[width=\linewidth]{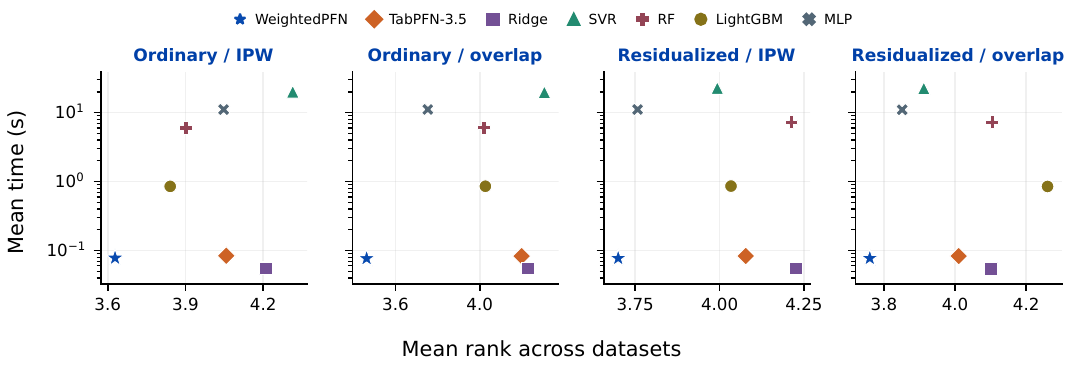}
\caption{{Mean policy-value rank and runtime (log scale) on 148 regression tasks. Lower values are better.}}
\label{fig:causal_reg}
\end{figure}

\begin{figure}[tbp]
\centering
\includegraphics[width=\linewidth]{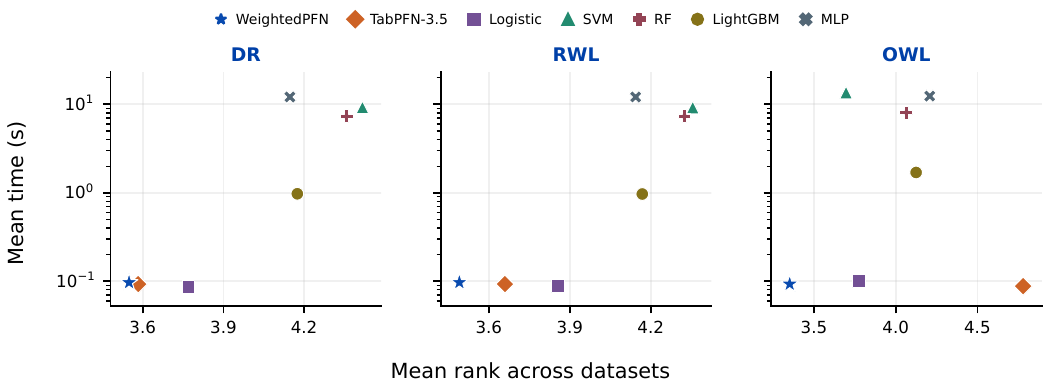}
\caption{{Mean policy-value rank and runtime (log scale) on 163 classification tasks each for DR and RWL and 84 for OWL. Lower values are better.}}
\label{fig:causal_clf}
\end{figure}

Figures~\ref{fig:causal_reg} and~\ref{fig:causal_clf} show the mean rank and mean runtime across tasks, and Table~\ref{tab:causal-performance} reports scaled policy values for regression and classification tasks.

Figures~\ref{fig:causal_reg}--\ref{fig:causal_clf} and Table~\ref{tab:causal-performance} show that WeightedPFN achieves the best mean rank and highest mean scaled policy value in all four weighted regression and three weighted classification settings. For regression, the largest gain occurs for the ordinary modified outcome with overlap weighting: the mean rank improves from 4.20 for TabPFN-3.5 to 3.46. For classification, the largest improvement occurs under OWL, where the mean rank improves from 4.78 for TabPFN-3.5 to 3.35. The gains over TabPFN-3.5 suggest that weighting helps recover the intended CATE or treatment rule. WeightedPFN also has runtimes comparable to those of the linear baselines and substantially lower than those of the nonlinear baselines. Its favorable performance and low runtime place it near the lower-left corner of Figures~\ref{fig:causal_reg} and~\ref{fig:causal_clf}. These results support its generalization from pretraining on synthetic missing-data tasks to real causal tasks.

\begin{table}[!htbp]
\centering \singlespacing
\caption{{Treatment selection performance (mean $\pm$ SD across tasks). Higher scaled policy values are better. Bold indicates the best mean in each column.}}\label{tab:causal-performance}
\fontsize{9}{11}\selectfont\setlength{\tabcolsep}{1.35pt}
\begin{tabular*}{\linewidth}{@{\extracolsep{\fill}}lcccc@{}}
\toprule
\multicolumn{5}{@{}l}{\textbf{Causal regression}}\\
 & \multicolumn{4}{c}{Scaled policy value (mean $\pm$ SD)}\\
\cmidrule(lr){2-5}
Method & \shortstack{Ordinary\\IPW} & \shortstack{Ordinary\\overlap} & \shortstack{Residualized\\IPW} & \shortstack{Residualized\\overlap}\\
\midrule
WeightedPFN & $\mathbf{1.296}\,{\pm}\,2.398$ & $\mathbf{1.298}\,{\pm}\,2.403$ & $\mathbf{1.310}\,{\pm}\,2.420$ & $\mathbf{1.311}\,{\pm}\,2.421$\\
\addlinespace[1.5pt]
TabPFN-3.5 & $1.280\,{\pm}\,2.401$ & $1.280\,{\pm}\,2.401$ & $1.301\,{\pm}\,2.424$ & $1.301\,{\pm}\,2.424$\\
\addlinespace[1.5pt]
Ridge & $1.279\,{\pm}\,2.407$ & $1.278\,{\pm}\,2.408$ & $1.302\,{\pm}\,2.421$ & $1.307\,{\pm}\,2.419$\\
SVR & $1.283\,{\pm}\,2.399$ & $1.280\,{\pm}\,2.403$ & $1.303\,{\pm}\,2.412$ & $1.305\,{\pm}\,2.413$\\
RF & $1.287\,{\pm}\,2.406$ & $1.283\,{\pm}\,2.413$ & $1.298\,{\pm}\,2.423$ & $1.297\,{\pm}\,2.418$\\
LightGBM & $1.286\,{\pm}\,2.398$ & $1.282\,{\pm}\,2.407$ & $1.294\,{\pm}\,2.415$ & $1.292\,{\pm}\,2.413$\\
MLP & $1.282\,{\pm}\,2.406$ & $1.290\,{\pm}\,2.401$ & $1.306\,{\pm}\,2.418$ & $1.302\,{\pm}\,2.417$\\
\bottomrule
\end{tabular*}\par\medskip
\begin{tabular*}{\linewidth}{@{\extracolsep{\fill}}lccc@{}}
\toprule
\multicolumn{4}{@{}l}{\textbf{Causal classification}}\\
 & \multicolumn{3}{c}{Scaled policy value (mean $\pm$ SD)}\\
\cmidrule(lr){2-4}
Method & DR & RWL & OWL\\
\midrule
WeightedPFN & $\mathbf{1.198}\,{\pm}\,2.484$ & $\mathbf{1.198}\,{\pm}\,2.485$ & $\mathbf{1.417}\,{\pm}\,2.548$\\
\addlinespace[1.5pt]
TabPFN-3.5 & $1.191\,{\pm}\,2.487$ & $1.190\,{\pm}\,2.488$ & $1.332\,{\pm}\,2.546$\\
\addlinespace[1.5pt]
Logistic & $1.194\,{\pm}\,2.483$ & $1.192\,{\pm}\,2.485$ & $1.386\,{\pm}\,2.554$\\
SVM & $1.180\,{\pm}\,2.487$ & $1.181\,{\pm}\,2.485$ & $1.392\,{\pm}\,2.543$\\
RF & $1.185\,{\pm}\,2.488$ & $1.188\,{\pm}\,2.484$ & $1.380\,{\pm}\,2.529$\\
LightGBM & $1.186\,{\pm}\,2.481$ & $1.185\,{\pm}\,2.479$ & $1.381\,{\pm}\,2.538$\\
MLP & $1.191\,{\pm}\,2.481$ & $1.188\,{\pm}\,2.481$ & $1.376\,{\pm}\,2.545$\\
\bottomrule
\end{tabular*}
\end{table}

\subsection{Ablation}
\label{sec:ablation}

This ablation study examines how fine-tuning on synthetic missing-data tasks and the use of sample weights when making predictions contribute to model performance.

For both regression and classification, we compare four variants: (i) TabPFN-3.5 without incorporating sample weights; (ii) TabPFN-3.5 with sample weights using weighted attention in Section~\ref{sec:model-architecture}; (iii) WeightedPFN without sample weights, so that weighted attention reduces to standard attention; and (iv) WeightedPFN. The two WeightedPFN variants use the same checkpoint after 800 fine-tuning updates.

Tables~\ref{tab:ablation-talent-settings} and~\ref{tab:ablation-causal-settings} report scaled RMSE and accuracy for missing-data tasks and scaled policy value for causal tasks.

On TALENT missing-data tasks, combining fine-tuning with sample weights yields the lowest mean scaled RMSE for regression and the highest mean accuracy for classification under X+Y and Y-only selection. In these settings, WeightedPFN, which uses both weighted attention and fine-tuning on synthetic missing-data tasks, outperforms both TabPFN-3.5 with sample weights and WeightedPFN without sample weights. Under X-only and MCAR selection, WeightedPFN performs slightly worse than unweighted TabPFN-3.5, and most of this gap is already present when WeightedPFN predicts without weights. This reflects the trade-off noted in Section~\ref{sec:exp-talent}: fine-tuning brings large gains under X+Y and Y-only selection at the cost of a small loss under X-only and MCAR selection. On causal tasks, our WeightedPFN using both weighted attention and fine-tuning achieves the highest mean scaled policy value in all seven settings. These results suggest that the benefits of fine-tuning with weighted attention extend to causal tasks outside the missing-data task family used for pretraining.

\begin{table}[!htbp]
\centering\singlespacing
\caption{TALENT ablation: scaled RMSE and accuracy (\%), reported as mean $\pm$ SD across datasets. Bold indicates the best mean in each setting.}\label{tab:ablation-talent-settings}
\fontsize{9}{11}\selectfont\setlength{\tabcolsep}{2pt}
\begin{tabular*}{\linewidth}{@{\extracolsep{\fill}}l*{4}{c}@{}}
\toprule
& \multicolumn{2}{c}{TabPFN-3.5} & \multicolumn{2}{c}{WeightedPFN}\\
\cmidrule(lr){2-3}\cmidrule(lr){4-5}
& Weights off & Weights on & Weights off & Weights on\\
\midrule
\multicolumn{5}{@{}l}{\textbf{Regression}}\\
Setting & Scaled RMSE & Scaled RMSE & Scaled RMSE & Scaled RMSE\\
\midrule
X+Y & $0.583\,{\pm}\,0.378$ & $0.580\,{\pm}\,0.358$ & $0.591\,{\pm}\,0.390$ & $\mathbf{0.547}\,{\pm}\,0.310$\\
Y-only & $0.585\,{\pm}\,0.367$ & $0.587\,{\pm}\,0.357$ & $0.583\,{\pm}\,0.356$ & $\mathbf{0.554}\,{\pm}\,0.316$\\
X-only & $\mathbf{0.467}\,{\pm}\,0.294$ & $0.470\,{\pm}\,0.296$ & $0.473\,{\pm}\,0.297$ & $0.474\,{\pm}\,0.298$\\
MCAR & $\mathbf{0.448}\,{\pm}\,0.289$ & $\mathbf{0.448}\,{\pm}\,0.289$ & $0.458\,{\pm}\,0.295$ & $0.458\,{\pm}\,0.295$\\
\midrule
\multicolumn{5}{@{}l}{\textbf{Classification}}\\
Setting & Accuracy (\%) & Accuracy (\%) & Accuracy (\%) & Accuracy (\%)\\
\midrule
X+Y & $75.74\,{\pm}\,18.48$ & $76.09\,{\pm}\,17.49$ & $75.66\,{\pm}\,18.60$ & $\mathbf{81.28}\,{\pm}\,14.93$\\
Y-only & $75.84\,{\pm}\,18.04$ & $75.58\,{\pm}\,17.26$ & $75.78\,{\pm}\,18.04$ & $\mathbf{79.33}\,{\pm}\,16.03$\\
X-only & $\mathbf{83.68}\,{\pm}\,13.71$ & $83.53\,{\pm}\,13.85$ & $83.57\,{\pm}\,13.74$ & $83.52\,{\pm}\,13.81$\\
MCAR & $\mathbf{84.27}\,{\pm}\,13.71$ & $\mathbf{84.27}\,{\pm}\,13.71$ & $84.18\,{\pm}\,13.74$ & $84.18\,{\pm}\,13.74$\\
\bottomrule
\end{tabular*}
\end{table}

\begin{table}[!htbp]
\centering\singlespacing
\caption{Causal ablation: scaled policy value (mean $\pm$ SD across tasks). Bold indicates the highest mean in each setting.}\label{tab:ablation-causal-settings}
\fontsize{9}{11}\selectfont\setlength{\tabcolsep}{2pt}
\begin{tabular*}{\linewidth}{@{\extracolsep{\fill}}l*{4}{c}@{}}
\toprule
& \multicolumn{2}{c}{TabPFN-3.5} & \multicolumn{2}{c}{WeightedPFN}\\
\cmidrule(lr){2-3}\cmidrule(lr){4-5}
& Weights off & Weights on & Weights off & Weights on\\
\midrule
Setting & Scaled value & Scaled value & Scaled value & Scaled value\\
\midrule
\multicolumn{5}{@{}l}{\textbf{Regression}}\\
Ordinary / IPW & $1.280\,{\pm}\,2.401$ & $1.280\,{\pm}\,2.401$ & $1.287\,{\pm}\,2.402$ & $\mathbf{1.296}\,{\pm}\,2.398$\\
Ordinary / overlap & $1.280\,{\pm}\,2.401$ & $1.278\,{\pm}\,2.401$ & $1.287\,{\pm}\,2.402$ & $\mathbf{1.298}\,{\pm}\,2.403$\\
Residualized / IPW & $1.301\,{\pm}\,2.424$ & $1.305\,{\pm}\,2.416$ & $1.308\,{\pm}\,2.418$ & $\mathbf{1.310}\,{\pm}\,2.420$\\
Residualized / overlap & $1.301\,{\pm}\,2.424$ & $1.310\,{\pm}\,2.415$ & $1.308\,{\pm}\,2.418$ & $\mathbf{1.311}\,{\pm}\,2.421$\\
\midrule
\multicolumn{5}{@{}l}{\textbf{Classification}}\\
DR & $1.191\,{\pm}\,2.487$ & $1.195\,{\pm}\,2.481$ & $1.190\,{\pm}\,2.487$ & $\mathbf{1.198}\,{\pm}\,2.484$\\
RWL & $1.190\,{\pm}\,2.488$ & $1.196\,{\pm}\,2.480$ & $1.191\,{\pm}\,2.487$ & $\mathbf{1.198}\,{\pm}\,2.485$\\
OWL & $1.332\,{\pm}\,2.546$ & $1.390\,{\pm}\,2.549$ & $1.334\,{\pm}\,2.539$ & $\mathbf{1.417}\,{\pm}\,2.548$\\
\bottomrule
\end{tabular*}
\end{table}

\FloatBarrier
{
\section{Discussion}
\label{sec:discussion}

In this paper, we propose a general pipeline that adapts attention-based tabular foundation models to weighted regression and classification by combining synthetic missing-data tasks, weighted attention, and a normalized weighted loss. Applied to TabPFN-3.5, it yields WeightedPFN after approximately half an hour of fine-tuning on a single H100 GPU. Across real missing-data and out-of-distribution causal tasks, WeightedPFN combines strong performance with runtimes close to those of linear baselines, without task-specific fine-tuning. Because the pipeline is not specific to TabPFN-3.5, it can also be applied to future attention-based tabular foundation models, carrying their advances over to weighted statistical learning.

% Future work could extend the adaptation pipeline to other weighted statistical problems, including survival analysis and time-series forecasting. Such extensions would require synthetic tasks that capture censoring or temporal dependence, architectures suited to these data structures, and appropriate weighted training losses. The aim would be to develop reusable foundation models that can use sample weights in these settings while retaining fast prediction without task-specific fine-tuning.

Our theory shows that, under regularity and task-coverage conditions, a suitably pretrained weighted Transformer attains the nonparametric minimax rate for weighted regression on new tasks without task-specific fitting. This result could be extended to weighted classification and to Transformer architectures that more closely match practical tabular foundation models. A related direction is to establish theoretical conditions under which pretraining on missing-data tasks supports generalization to other weighted problems, such as those arising in causal inference. We leave these extensions to future work.
}

\section*{Data availability statement}
The TALENT datasets used in Section~\ref{sec:exp-talent} are publicly available \citep{liu2025talent}. The randomized-experiment datasets used in Section~\ref{sec:exp-causal}, most of which are from RCT\_Bench \citep{shao2026covariates}, are publicly available and are listed in Section~\ref{s:causal-datasets} of the supplementary material. The fine-tuned weights of WeightedPFN are available at \url{https://huggingface.co/xiajunwen/WeightedPFN}, and the inference code is available at \url{https://github.com/xiajunwen1007/WeightedPFN}.

\section*{Acknowledgments}
{Generative AI tools were used for English polishing and coding assistance.}

\bibliographystyle{apalike}
\bibliography{references}

\end{document}